\documentclass[runningheads,a4paper]{llncs}
\usepackage[cmex10]{amsmath}
\usepackage{amssymb}
\usepackage{llncsdoc}
\usepackage{mathptmx}       
\usepackage{helvet}         
\usepackage{courier}        
\usepackage{makeidx}         
\usepackage[pdftex]{graphicx}
\usepackage{epstopdf}
\DeclareGraphicsExtensions{.pdf,.jpeg,.png,.eps}
\usepackage{multirow}        
\usepackage[bottom]{footmisc}

\usepackage{url}
\usepackage[linesnumbered,ruled,vlined]{algorithm2e}
\usepackage{array}
\usepackage{booktabs}
\usepackage[caption=false,font=footnotesize]{subfig}
\usepackage{xcolor,colortbl}
\usepackage{bm}
\usepackage{mathtools}

\DeclareMathSymbol{\R}{\mathalpha}{AMSb}{"52}

\definecolor{Gray}{gray}{0.85}
\definecolor{LightCyan}{rgb}{0.88,1,1}

\newcolumntype{P}[1]{>{\centering\arraybackslash}p{#1}}

\begin{document}
\mainmatter 

\title{Efficient data augmentation using graph imputation neural networks}

\author{Indro Spinelli \and Simone Scardapane \and Michele Scarpiniti \and Aurelio Uncini}
\titlerunning{Data imputation using GINNs}
\authorrunning{Spinelli et al.}
\institute{Department of Information Engineering, Electronics and Telecommunications (DIET), \\ ``Sapienza'' University of Rome, \\ Via Eudossiana 18, 00184, Rome. \\Email: \{simone.scardapane; michele.scarpiniti; aurelio.uncini\}@uniroma1.it}

\maketitle

\begin{abstract}
Recently, data augmentation in the semi-supervised regime, where unlabeled data vastly outnumbers labeled data, has received a considerable attention. In this paper, we describe an efficient technique for this task, exploiting a recent framework we proposed for missing data imputation called graph imputation neural network (GINN). The key idea is to leverage both supervised and unsupervised data to build a graph of similarities between points in the dataset. Then, we augment the dataset by severely damaging a few of the nodes (up to 80\% of their features), and reconstructing them using a variation of GINN. On several benchmark datasets, we show that our method can obtain significant improvements compared to a fully-supervised model, and we are able to augment the datasets up to a factor of 10x. This points to the power of graph-based neural networks to represent structural affinities in the samples for tasks of data reconstruction and augmentation.
\end{abstract}

\section{Introduction}
\label{sec:intro}

Semi-supervised learning (SSL) studies how to exploit vast amounts of unlabeled data to improve the performances of a model trained on a smaller number of labeled data points \cite{chapelle2010semisupervised}. Over the years, a large number of solutions were devised, ranging from manifold regularization \cite{belkin2006manifold} to label propagation \cite{grandvalet2005semi}. With the emergence of deep learning techiques and the increase in datasets' size, several new methods were also proposed to leverage end-to-end training of deep networks in the SSL regime, e.g., pseudo-labels \cite{lee2013pseudo} and ladder networks \cite{antii2015semisupervised}.

An interesting line of research recently concerns the exploitation of unsupervised information to perform data augmentation \cite{cui2015data}. Augmented datasets can then be used directly, or indirectly through the imposition of one or more regularizers on the learning process, by enforcing the model to have stable outputs with respect to the new points \cite{grandvalet2005semi}. Several papers have shown that semi-supervised data augmentation has the potential to provide significant boosts in accuracy. for deep learning models compared to standard SSL algorithms or equivalent fully-supervised solutions \cite{2019arXiv190412848X,berthelot2019mixmatch}.

In this light, an essential research question concerns how to devise efficient and general-purpose strategies for data augmentation. A popular data augmentation strategy, mixup \cite{zhang2017mixup}, builds new datapoints by taking linear combinations of points in the dataset, but it requires knowledge of the underlying labels and has shown mixed results when dealing with non-image data. Other popular techniques, such as AutoAugment \cite{cubuk2018autoaugment}, are instead especially designed to work in structured domains such as images or audio samples.

In this paper, we propose a new method to perform data augmentation from general vectorial inputs. Our starting point is the graph imputer neural network (GINN) \cite{spinelli2019ginn}, an algorithm we proposed recently for multivariate data imputation in the presence of missing features. GINN works by building a graph of similarities from the points in the dataset (similarly to manifold regularization \cite{belkin2006manifold}). Missing features are then estimated by applying a customized graph autoencoder \cite{kipf2016semi} to reconstruct the full dataset starting from the incomplete dataset and the graph information \cite{spinelli2019ginn}. GINN has shown to outperform most state-of-the-art methods for data imputation, leading us to investigate whether its performance can be extended to the case of data augmentation.

The algorithm proposed in this paper is an application of GINN showing its effectiveness for data augmentation. After constructing the similarity graph starting from both labeled and unlabeled data, we corrupt some of the nodes by removing up to $80\%$ of their features. After recomputing on-the-fly their connections with the neighboring nodes, we apply the previously trained GINN architecture for performing imputation, effectively generating a new data point that can be added to the dataset. Despite its conceptual simplicity, we show through an extensive experimental evaluation that augmenting the dataset in this fashion (up to 5/10 times) leads to significant improvements in accuracy when training standard supervised learning algorithms.

The rest of the paper is structured as follows. In Section \ref{sec:semi_supservised_data_augmentation_with_ginn} we describe our proposed technique for data augmentation. Section \ref{sec:experimental_evaluation} shows an experimental evaluation on a range of different datasets. We conclude with a few open research directions in Section \ref{sec:conclusions}.

\section{Semi-supervised data augmentation with GINN}
\label{sec:semi_supservised_data_augmentation_with_ginn}

\begin{figure}
    \centering
    \includegraphics[scale=0.7]{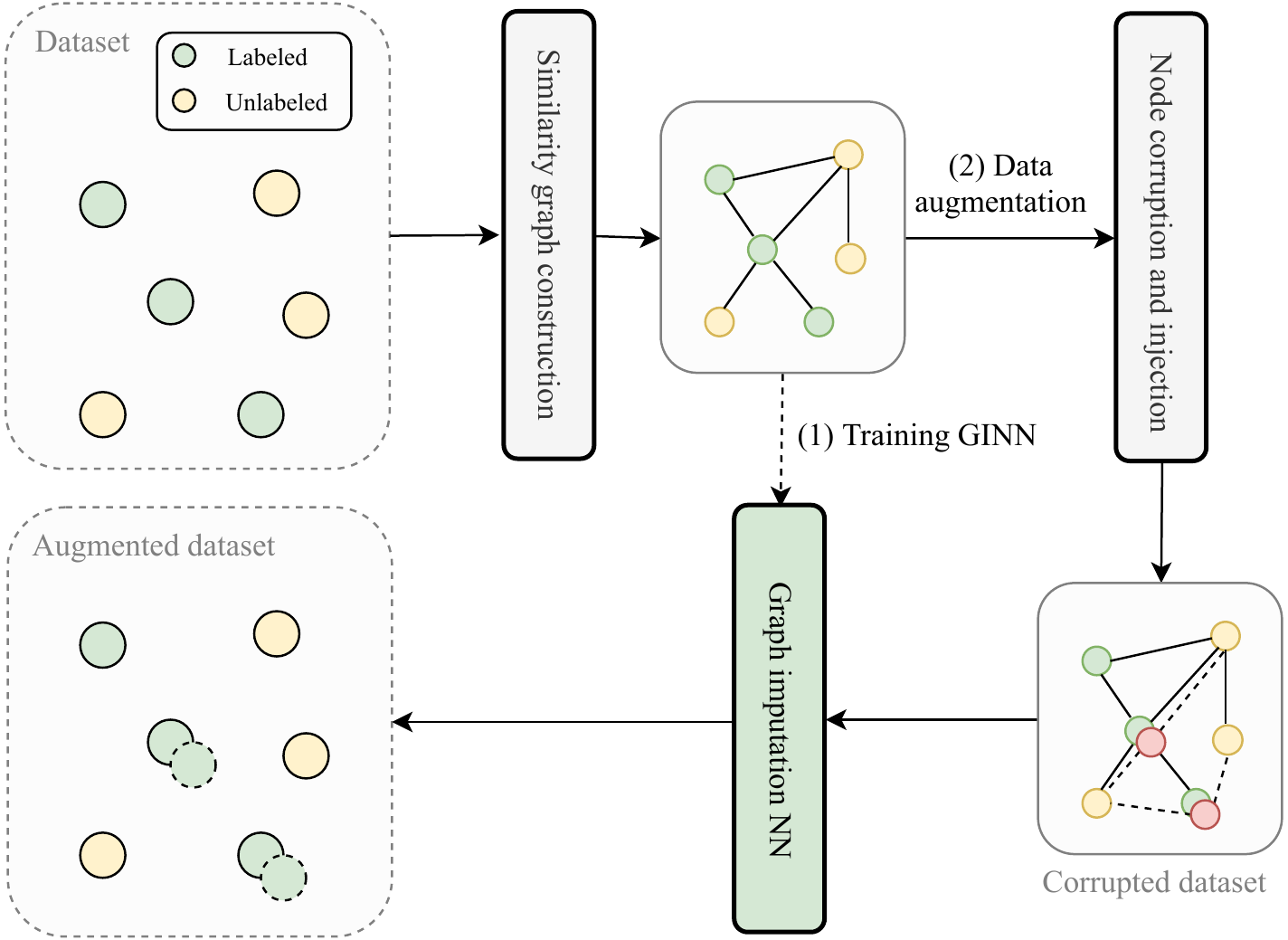}
    \caption{Overall schema of the proposed framework for data augmentation. In green we show the GINN method which is optimized from the dataset.}
    \label{fig:framework}
\end{figure}

Fig. \ref{fig:framework} shows a high-level overview of our framework. As we stated before, we first build a similarity graph from the data, used to train a GINN model \cite{spinelli2019ginn}. Then, we employ GINN on highly-corrupted points from the dataset to generate new datapoints. We first describe GINN in Section \ref{subsec:ginn}, before describing how we use it for data augmentation in Section \ref{subsec:semisupervised_data-augmentation}.

\subsection{Graph imputer neural network (GINN)}
\label{subsec:ginn}

GINN \cite{spinelli2019ginn} is a graph-based neural network that takes a damaged dataset as input (i.e., a dataset with a few missing entries), and is able to reconstruct the missing features by constructing a similarity graph and exploiting the graph convolution operation \cite{kipf2016semi,battaglia2018relational}. Here we summarize the GINN architecture in broad terms, while for a more in-depth description we refer the interested reader to \cite{spinelli2019ginn}.

Consider a generic dataset $\mathbf{X} \in \mathbb{R}^{n \times d}$, where each row encodes one example, defined by a vector of $d$ features. We would like a model that can reconstruct the original $\mathbf{X}$ even when a few of its elements are missing, i.e., an algorithm performing missing data imputation.

In order to train GINN to this end, we first augment the matrix information in $\mathbf{X}$ with a graph describing the structural proximity between points. In particular, we encode each feature vector as a node in a graph $\mathbf{G}$. The adjacency matrix $\mathbf{A}$ of the graph is derived from a similarity matrix $\mathbf{S}$ of the feature vectors, containing the pairwise Euclidean distances of the feature vectors. In order to keep only the most relevant edges, we apply a two-step pruning on $\mathbf{S}$. We compute the $97.72nd$ percentile for each row, corresponding to $+2\sigma$ under a Gaussian assumption, and we use this as a threshold, discarding all the connection below this value. The second step replicates the first one but acts on all the surviving elements of the matrix $\mathbf{S}$ (see the original paper \cite{spinelli2019ginn} for a rationale of this technique).

The core of GINN is a graph-convolutional autoencoder described as follows: 

\begin{align}
\mathbf{H} & = \text{ReLU} \left( \mathbf{L} \mathbf{X} \bm{\Theta}_1 \right) \,, \label{eq:ginn1}\\
\mathbf{\widehat X} & = \text{Sigmoid}\left( \mathbf{L} \mathbf{H} \bm{\Theta}_2 +  \widetilde{\mathbf{L}} \mathbf{X} \bm{\Theta}_3  + \bm{\Theta}_4\mathbf{g} \right ) \,. \label{eq:ginn2}
\end{align}

\noindent The graph encoder in Eq. \eqref{eq:ginn1}, maps  the input $\mathbf{X}$ to an intermediate representation $\mathbf{H}$ in an higher dimensional space, using a graph convolutional operation \cite{kipf2016semi}, where $\mathbf{L}$ is the Laplacian matrix associated to the graph and $\bm{\Theta}_1$ is a matrix of adaptable coefficients. The decoder in Eq. \eqref{eq:ginn2} maps the intermediate representation back to the original dimensional space providing a reconstructed dataset $\mathbf{\widehat X}$ with no missing values. As can be seen in Eq. \eqref{eq:ginn2}, we have two additional terms in the reconstruction process:

\begin{enumerate}
    \item The first is a skip connection with parameters $\bm{\Theta}_2$, which is always a graph convolution that propagates the information across the immediate neighbors of a node without the node itself. $ \widetilde{\mathbf{L}}$ is thus derived from the adjacency matrix $\mathbf{A}$ without self-loops in order to weight more the contribution of 1-hop nodes and to prevent the autoencoder to learn the identity function.
    \item The second additional term models the global properties of the graph, whose inclusion and properties have been described in \cite{battaglia2018relational}, with trainable parameters $\bm{\Theta}_3$. We introduce a global attribute vector $\mathbf{g}$ for the graph that contains statistical information of the dataset including the mean or mode of every attribute.
\end{enumerate} 

\noindent Both steps are described in more depth in \cite{spinelli2019ginn}. The network is trained to reconstruct the original dataset by minimizing the sum of three terms:

\begin{align}
L_{A} & = \alpha \text{MSE}(\mathbf{X}, \widehat{\mathbf{X}}) + (1 - \alpha) \text{CE}(\mathbf{X}, \widehat{\mathbf{X}}) + \gamma \,\text{MSE}(\text{global}(\mathbf{\widehat{X}}), \text{global}(\mathbf{X})) \,. \label{eq:ginn3}
\end{align}

\noindent The loss function in Eq. \eqref{eq:ginn3} minimizes the reconstruction error over the non-missing elements combining the mean squared error (MSE) for the numerical variables and the cross-entropy (CE) for the categorical variables. In addition, since the computation of the global information is differentiable, we compute a loss term  with respect to the global attributes of the original dataset. Here, $\alpha$ and $\gamma$ are additional hyperparameters, the first is initialized as the ratio between numerical and categorical columns, the second is a weighting factor. Practically, at every iteration of optimization we randomly drop a given percentage of elements in $\mathbf{X}$ in order to learn to reconstruct the original matrix irrespective of which elements are missing.

This model can be trained as standalone or paired with a critic network $C$ in an adversarial fashion \cite{yoon2018gain}. In the latter case we have a 3-layer feed-forward network that learns to distinguish between imputed ($\hat{\mathbf{x}} \sim \mathbb{P}_{imp}$) and real ($\mathbf{x} \sim \mathbb{P}_{real}$) data. To train the models together we use the Wasserstein distance \cite{arjovsky2017}. The objective function of the critic is:
\begin{equation}
\min _{A} \max _{C \in \mathcal{D}} \underset{\mathbf{x} \sim \mathbb{P}_{real}}{\mathbb{E}}[C(\mathbf{x})]-\underset{\hat{\mathbf{x}} \sim \mathbb{P}_{imp}}{\mathbb{E}}[C(\hat{\mathbf{x}})) ] \,,
\label{eq:wasserstein_loss}
\end{equation}
\noindent where $\mathcal{D}$ is the set of 1-Lipschitz functions, $\mathbb{P}_{imp}$  is the model distribution implicitly defined by our GCN autoencoder $A$, and  $\mathbb{P}_{real}$ is the unknown data distribution.

In addition, we use a gradient penalty introduced in \cite{gulrajani2017} to enhance training stability and enforcing the Lipschitz constraint, obtaining the final critic loss:
\begin{equation}
   L_{C}=\underset{\hat{\mathbf{x}} \sim \mathbb{P}_{imp}}{\mathbb{E}}[C(\hat{\mathbf{x}})]-\underset{\mathbf{x} \sim \mathbb{P}_{real}}{\mathbb{E}}[C(\mathbf{x})]+\lambda \underset{\tilde{\mathbf{x}} \sim \mathbb{P}_{\tilde{\mathbf{x}}}}{\mathbb{E}}\left[\left(\left\|\nabla_{\tilde{\mathbf{x}}} C(\tilde{\mathbf{x}})\right\|_{2}-1\right)^{2}\right] \,,
\end{equation}
where $\lambda$ is an additional hyper-parameter.
We define $\mathbb{P}_{\tilde{\mathbf{x}}}$ as sampling uniformly from the combination of the real distribution $\mathbb{P}_{imp}$ and from the distribution resulting from the imputation $\mathbb{P}_{imp}$. This means that the feature vector $\tilde{\mathbf{x}}$ will be composed by both real and imputed elements in almost equal size.

The total loss of the autoencoder becomes:
\begin{equation}
L_T = L_A - \underset{\hat{\mathbf{x}} \sim \mathbb{P}_{imp}}{\mathbb{E}}[C(\hat{\mathbf{x}})] \,,
\end{equation}
since it must fool the critic and minimize the reconstruction error at the same time.
In our implementation the GCN autoencoder is trained once for every five optimization steps of the critic and they will influence each other during the whole process.

\subsection{Semi-supervised data augmentation}
\label{subsec:semisupervised_data-augmentation}

In order to augment the dataset in the semi-supervised setting, we build the graph with labeled and unlabeled features and let GINN learn its representation. Rather than learning the labels, like in a transductive setting \cite{kipf2016semi}, we want to generate completely new nodes in the graph and thus new feature vectors with labels that can be used later for other objectives.
 To do so we take all the labeled data and use it to generate a new data matrix of the size of the desired augmentation level. This means that features vectors can be repeated in this matrix. We then damage the matrix in a MCAR (Missing Completely At Random) fashion removing 80\% of its elements. These severely damaged vectors will be the starting point of our data augmentation strategy. We note that in \cite{spinelli2019ginn} we  have shown already the ability of GINN  in imputing vectors having few non-zero elements.
 
We inject these new damaged feature vectors in the graph. To recompute their connections on-the-fly, we follow the same procedure described above, without considering the second pruning step in order to guarantee that every node will have at least one neighbor. The connections will be computed only with unlabeled nodes in order to prevent the new nodes to be too similar to the ones they originated from. This time the similarity has to take into account the fact that the new vectors will have only few non-zero elements. For this reason we formulate the distance as follows:

\begin{equation}
S_{ij}  = d(\mathbf{x}_i \odot ( \mathbf{m}_i \odot \mathbf{m}_j), \mathbf{x}_j \odot (\mathbf{m}_i \odot \mathbf{m}_j)) \,,
\end{equation}
where $\odot$ stands for the Hadamard product between vectors, $\mathbf{m}_i$ and $\mathbf{m}_j$ are binary vectors describing the missing elements in $\mathbf{x}_i$ and $\mathbf{x}_j$ respectively, and $d$ is the Euclidean distance.

Once the new data is in the graph, we use GINN to impute all missing elements, and add the resulting vectors to our original dataset. The resulting imputed vectors will have the label of the node they have been generated from and will be influenced by the unlabeled nodes in the graph sharing similar features.

\section{Experimental evaluation}
\label{sec:experimental_evaluation}

In this section, we analyze the data augmentation performance of our framework.
For the evaluation we used 6 classification datasets from the UCI Machine Learning Repository,\footnote{\url{http://archive.ics.uci.edu/ml/index.php}} and their characteristics are described in Table \ref{tab:datasets}. These datasets contain  numerical, categorical and mixed features in order to show that our framework is capable of generating realistic feature vectors composed of different types of attributes.

\begin{table*}[!ht]
\centering
\caption{Datasets used in our experiments. All datasets  were downloaded from the UCI repository.}\vspace{0.5em}
\begin{tabular}{l|c|c|c}
Name & observations & numerical attr.    & categorical attr.\\
\midrule
abalone                    & 4177  & 8  & 0  \\
heart                      & 303   & 8  & 5  \\
ionosphere                 & 351   & 34 & 0  \\
phishing                   & 1353  & 0  & 9  \\
tic-tac-toe                & 958   & 9  & 9  \\
wine-quality-red           & 1599  & 11 & 0  \\
\bottomrule
\end{tabular}%
\label{tab:datasets}
\end{table*}

We tracked the performances of 5 different classifiers when using the default and the augmented datasets. The algorithms used are a k-NN classifier with $k=5$, regularized logistic regression (LOG), C-Support Vector Classification with an RBF kernel (SVC), a random forest classifier with 10 estimators (RF) and a maximum depth of 10 and a feedforward neural network composed of six layers with an embedding dimension of 128 (MLP). All hyper-parameters are initialized with their default values in the scikit-learn implementation.

Concerning our framework, we use an embedding dimension of the hidden layer of $128$, sufficient for an overcomplete representation for all the datasets involved. We trained GINN for a maximum of $3000$ epochs with an early stopping strategy for the reconstruction loss over the known elements. The critic used is a simple 3-layer feed-forward network trained 5 times for each optimization step of the autoencoder. We used the Adam optimizer \cite{kingma2014adam} for both networks with a learning rate of $1\times10^{-3}$ and $1\times10^{-5}$ respectively for autoencoder and critic. All other hyper-parameters are taken from the original GINN paper \cite{spinelli2019ginn}, whose implementation is available as an open-source package.\footnote{\url{https://github.com/spindro/GINN}}

\subsection{Semi-Supervised Classification}

In the semi-supervised setting, we divided our data between the training set, 70\%,  and test set, 30\%. Only 10\% of the training set has labels associated with feature vectors. We augment the dataset with GINN creating 3 different versions of the training set, respectively having $2x$, $5x$ and $10x$ more labeled data. We then train the classifiers on this 4 different training sets and compute the accuracy, averaged over 5 different trials. Each trial has different splits of training and test sets. 

\begin{table*}[!ht]
\centering
\caption{Mean classification accuracy over 5 different trials on different splits of data obtained using the standard dataset  \textbf{X} and the augmented versions with GINN. In Figure \ref{fig:summ} we summarize the results of this table.}\vspace{0.5em}
\resizebox{0.8\textwidth}{!}{%
\begin{tabular}{p{0.15\textwidth}p{0.1\textwidth}P{0.15\textwidth}P{0.15\textwidth}P{0.15\textwidth}P{0.15\textwidth}P{0.15\textwidth}}
\textbf{Dataset} & \textbf{Classifier} & \textbf{Baseline} & \textbf{Augmented (2x)} & \textbf{Augmented (5x)} & \textbf{Augmented (10x)}\\
\midrule
  & LOG & 52.87 & 52.54 & 52.47 & 54.50 \\
 & k-NN & 52.07 & 52.07 & 52.07 & 52.07 \\
abalone & SVC & 52.87 & 52.87 & 52.87 & 52.87 \\
 & RF & 51.53 & 51.18 & 52.38 & 53.67 \\
 & MLP & 50.53 & 52.66 & 52.03 & 54.78 \\
 \midrule
 & LOG & 76.92 & 70.77 & 70.77 & 66.59 \\
 & k-NN & 58.24 & 64.40 & 64.40 & 62.56 \\
heart & SVC & 55.88 & 56.04 & 56.04 & 56.04 \\
 & RF & 79.56 & 81.10 & 80.44 & 78.02 \\
 & MLP & 65.71 & 66.81 & 63.96 & 61.32 \\
 \midrule
 & LOG & 78.30 & 80.94 & 79.06 & 78.49 \\
 & k-NN & 66.04 & 90.57 & 90.57 & 90.57 \\
ionosphere & SVC & 64.15 & 85.09 & 84.34 & 85.28 \\
 & RF & 83.96 & 87.92 & 85.47 & 86.23 \\
 & MLP & 90.57 & 88.87 & 86.04 & 86.04 \\
 \midrule
 & LOG & 83.50 & 82.07 & 82.02 & 81.67 \\
 & k-NN & 82.76 & 84.04 & 83.55 & 84.19 \\
phishing & SVC & 82.51 & 82.17 & 81.92 & 81.92 \\
 & RF & 81.33 & 81.58 & 81.08 & 81.82 \\
 & MLP & 81.38 & 81.97 & 83.05 & 82.91 \\
 \midrule
 & LOG & 77.43 & 73.89 & 69.86 & 70.07 \\
 & k-NN & 73.61 & 74.10 & 75.00 & 78.13 \\
tic-tac-toe & SVC & 65.28 & 64.65 & 67.85 & 68.47 \\
 & RF & 71.94 & 75.76 & 72.36 & 74.38 \\
 & MLP & 78.40 & 73.06 & 78.75 & 82.01 \\
 \midrule
 & LOG & 53.75 & 53.25 & 53.96 & 54.21 \\
 & k-NN & 45.83 & 46.21 & 49.67 & 49.79 \\
wine-quality & SVC & 45.63 & 47.67 & 52.42 & 52.38 \\
 & RF & 50.58 & 50.92 & 54.96 & 54.21 \\
 & MLP & 48.83 & 50.83 & 51.04 & 50.38 \\
\bottomrule
\end{tabular}%
}
\label{tab:data}
\end{table*}

\begin{figure}[!t]
    \centering
    \subfloat{\includegraphics[scale=0.8]{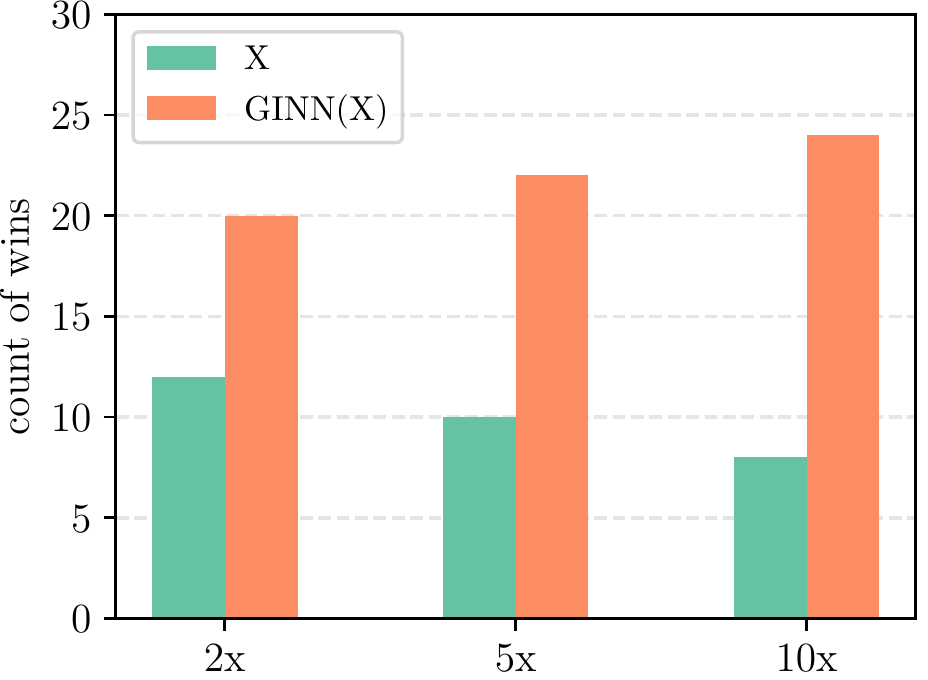}}
    \caption{Number of times the default and the augmented datasets with GINN had a better classification performances over 5 different trials considering all datasets and classifiers in the benchmark.}
    \label{fig:summ}
\end{figure}

In Table \ref{tab:data} we show the classification results. As can be seen, our method consistently improves classification accuracy. These improvements range from less than a percentage point up to an increment of 24 percentage points. When we do not improve the results, our decrease in performance regards at most 6 percentage points for a single classifier (e.g., LOG in the hearth dataset). In Figure \ref{fig:summ} we summarize the results of Table \ref{tab:data} and propose a comparison of the times our augmented datasets allows to achieve a better classification accuracy to the classifiers with respect to the original dataset. As can be clearly seen this difference increases with the size of the augmentation, showing that our framework is capable of generating good feature vectors even when creating a data matrix 10 times the size of the original dataset.

\section{Conclusions}
\label{sec:conclusions}

In this paper, we proposed an approach to data augmentation in the semi-supervised regime by reformulating it as a problem of extreme missing data imputation. To this end, we employ a novel algorithm for missing data imputation built on top of a graph autoencoder. Our results on a set of standard vectorial benchmarks show that the method can significantly improve over using only the labeled information, even when the dataset is augmented up to ten times its original size.

The method lends itself to a variety of improvements. First of all, we are interested in evaluating the augmentation strategies not only by directly retraining a supervised algorithms, but also in the context of several regularization strategies commonly used today \cite{2019arXiv190412848X}. We would also like to test the algorithm on non-vectorial domains, including images and audio, where the challenge is to define a proper metric to build the similarity graph. As a final remark, we note that the experiments presented here open the way to a set of interesting additional questions. In particular, viewing data augmentation as an extreme case of data imputation bridges two different fields with high potential for cross-fertilization.

\bibliographystyle{splncs03}
\bibliography{biblio}

\begin{thebibliography}{10}
\providecommand{\url}[1]{\texttt{#1}}
\providecommand{\urlprefix}{URL }

\bibitem{arjovsky2017}
Arjovsky, M., Chintala, S., Bottou, L.: {W}asserstein generative adversarial
  networks. In: Proc. 34th International Conference on Machine Learning (ICML).
  vol.~70, pp. 214--223 (2017)

\bibitem{battaglia2018relational}
Battaglia, P.W., Hamrick, J.B., Bapst, V., Sanchez-Gonzalez, A., Zambaldi, V.,
  Malinowski, M., Tacchetti, A., Raposo, D., Santoro, A., Faulkner, R., et~al.:
  Relational inductive biases, deep learning, and graph networks. arXiv
  preprint arXiv:1806.01261  (2018)

\bibitem{belkin2006manifold}
Belkin, M., Niyogi, P., Sindhwani, V.: Manifold regularization: A geometric
  framework for learning from labeled and unlabeled examples. Journal of
  Machine Learning Research  7(Nov),  2399--2434 (2006)

\bibitem{berthelot2019mixmatch}
Berthelot, D., Carlini, N., Goodfellow, I., Papernot, N., Oliver, A., Raffel,
  C.: Mixmatch: A holistic approach to semi-supervised learning. arXiv preprint
  arXiv:1905.02249  (2019)

\bibitem{chapelle2010semisupervised}
Chapelle, O., Schlkopf, B., Zien, A.: Semi-Supervised Learning. The MIT Press,
  1st edn. (2010)

\bibitem{cubuk2018autoaugment}
Cubuk, E.D., Zoph, B., Mane, D., Vasudevan, V., Le, Q.V.: Autoaugment: Learning
  augmentation policies from data. arXiv preprint arXiv:1805.09501  (2018)

\bibitem{cui2015data}
Cui, X., Goel, V., Kingsbury, B.: Data augmentation for deep neural network
  acoustic modeling. IEEE/ACM Transactions on Audio, Speech and Language
  Processing  23(9),  1469--1477 (2015)

\bibitem{grandvalet2005semi}
Grandvalet, Y., Bengio, Y.: Semi-supervised learning by entropy minimization.
  In: Advances in Neural Information Processing Systems. pp. 529--536 (2005)

\bibitem{gulrajani2017}
Gulrajani, I., Ahmed, F., Arjovsky, M., Dumoulin, V., Courville, A.C.: Improved
  training of wasserstein gans. In: Advances in Neural Information Processing
  Systems, pp. 5767--5777 (2017)

\bibitem{kingma2014adam}
Kingma, D.P., Ba, J.: Adam: A method for stochastic optimization. In: Proc. 3rd
  International Conference for Learning Representations (ICLR) (2014)

\bibitem{kipf2016semi}
Kipf, T.N., Welling, M.: Semi-supervised classification with graph
  convolutional networks. In: Proc. 2017 International Conference on Learning
  Representations (ICLR) (2017)

\bibitem{lee2013pseudo}
Lee, D.H.: Pseudo-label: The simple and efficient semi-supervised learning
  method for deep neural networks. In: Workshop on Challenges in Representation
  Learning, ICML. vol.~3, p.~2 (2013)

\bibitem{antii2015semisupervised}
Rasmus, A., Berglund, M., Honkala, M., Valpola, H., Raiko, T.: Semi-supervised
  learning with ladder networks. In: Advances in Neural Information Processing
  Systems, pp. 3546--3554 (2015)

\bibitem{spinelli2019ginn}
{Spinelli}, I., {Scardapane}, S., {Uncini}, A.: {Missing Data Imputation with
  Adversarially-trained Graph Convolutional Networks}. arXiv preprint
  arXiv:1905.01907  (2019)

\bibitem{2019arXiv190412848X}
{Xie}, Q., {Dai}, Z., {Hovy}, E., {Luong}, M.T., {Le}, Q.V.: {Unsupervised Data
  Augmentation}. arXiv preprint arXiv:1904.12848  (Apr 2019)

\bibitem{yoon2018gain}
Yoon, J., Jordon, J., van~der Schaar, M.: Gain: Missing data imputation using
  generative adversarial nets. In: Proc. 35th International Conference of
  Machine Learning (ICML). pp. 1--10 (2018)

\bibitem{zhang2017mixup}
Zhang, H., Cisse, M., Dauphin, Y.N., Lopez-Paz, D.: mixup: Beyond empirical
  risk minimization. arXiv preprint arXiv:1710.09412  (2017)

\end{thebibliography}

\end{document}